\documentclass[10pt,twocolumn,letterpaper]{article}

\usepackage[pagenumbers]{cvpr} % To force page numbers, e.g. for an arXiv version

\usepackage[dvipsnames]{xcolor}
\usepackage{graphics} % for pdf, bitmapped graphics files
\usepackage{epsfig} % for postscript graphics files
\usepackage{times} % assumes new font selection scheme installed
\usepackage{amsmath} % assumes amsmath package installed
\usepackage{amssymb}  % assumes amsmath package installed
\usepackage{enumitem}
\usepackage{mathtools}
\usepackage{algorithm}
\usepackage{algorithmic}
\usepackage{wrapfig}
\usepackage{colortbl}
\usepackage{listings}
\usepackage{afterpage}
\usepackage{relsize}
\usepackage{multicol}
\usepackage{multirow}
\usepackage{wasysym}
\usepackage{booktabs}
\usepackage{caption}
\usepackage{subcaption}
\usepackage{sidecap}
\usepackage{pifont}
\usepackage[nodisplayskipstretch]{setspace}
\usepackage{xcolor}
\usepackage{url}

\newcommand\mypar[1]{\par\vspace{-0.2mm}\noindent\textbf{#1}\;\;}
\definecolor{LightGrey}{rgb}{0.92,0.92,0.92}
\definecolor{Myred}{rgb}{1.00,0.12,0.36}
\definecolor{Myblue}{rgb}{0,0.60,0.87}
\newcommand{\ourwork}{SIG3D\xspace}

\definecolor{cvprblue}{rgb}{0.21,0.49,0.74}
\usepackage[pagebackref,breaklinks,colorlinks,citecolor=cvprblue]{hyperref}

\def\paperID{} % *** Enter the Paper ID here
\def\confName{CVPR}
\def\confYear{2024}

\title{Situational Awareness Matters in 3D Vision Language Reasoning
}

\author{Yunze Man \quad\quad\ Liang-Yan Gui \quad\quad Yu-Xiong Wang\\
University of Illinois Urbana-Champaign\\
{\tt\small \{yunzem2,lgui,yxw\}@illinois.edu}
}

\begin{document}

\maketitle

\begin{abstract}
\vspace{-0mm}
Being able to carry out complicated vision language reasoning tasks in 3D space represents a significant milestone in developing household robots and human-centered embodied AI. In this work, we demonstrate that a critical and distinct challenge in 3D vision language reasoning is situational awareness, which incorporates two key components: (1) The autonomous agent grounds its self-location based on a language prompt. (2) The agent answers open-ended questions from the perspective of its calculated position. To address this challenge, we introduce \ourwork, an end-to-end Situation-Grounded model for 3D vision language reasoning. We tokenize the 3D scene into sparse voxel representation and propose a language-grounded situation estimator, followed by a situated question answering module. Experiments on the SQA3D and ScanQA datasets show that \ourwork outperforms state-of-the-art models in situation estimation and question answering by a large margin (e.g., an enhancement of over 30\% on situation estimation accuracy). Subsequent analysis corroborates our architectural design choices, explores the distinct functions of visual and textual tokens, and highlights the importance of situational awareness in the domain of 3D question answering. The project page is available at \url{https://yunzeman.github.io/situation3d}.
\end{abstract}    
\section{Introduction}
\label{sec:introduction}

Humans learn knowledge efficiently through the interactions with the 3D world and the integration of multi-modal information, such as verbal guidance or instructions. Similarly, introducing language guidance into the visual comprehension task can greatly enhance the learning efficiency of models~\cite{alayrac2022flamingo,lin2023multimodality}. Nonetheless, despite considerable advancements in linguistic understanding~\cite{kenton2019bert,brown2020language,touvron2023llama,chowdhery2022palm} and vision-language integration~\cite{alayrac2022flamingo,li2022blip,radford2021learning,wang2023visionllm}, current methodologies remain deficient in accurately perceiving and rationalizing within real-world 3D environments, which is largely attributed to the lack of 3D situational reasoning capabilities.

\begin{figure}
    \centering
    \captionsetup{type=figure}
    \includegraphics[trim=270 220 260 220, clip=True, width=1\linewidth]{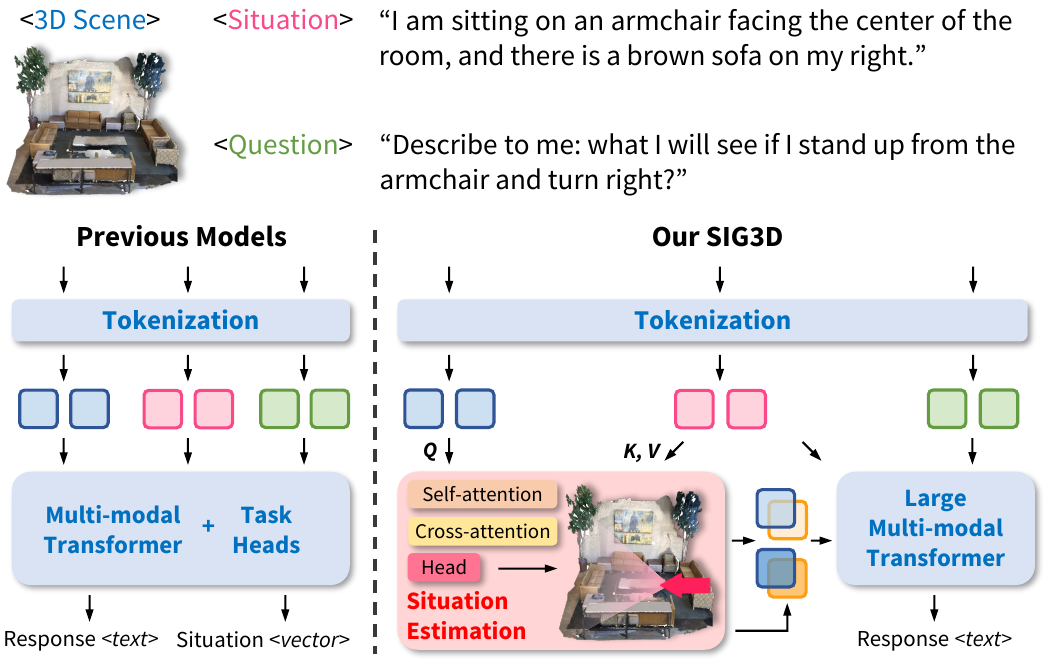}
    \vspace{-6mm}
    \captionof{figure}{Previous methods perform direct 3D vision language reasoning without modeling the situation of an embodied agent in the 3D environment. Our method, \ourwork, grounds the situational description in the 3D space, and then re-encodes the visual tokens from the agent's intended perspective before vision-language fusion, resulting in a more comprehensive and generalized 3D vision language (VL) representation and reasoning framework. Q, K, V stand for query, key, and value, respectively.
    } 
    \vspace{-4mm}
    \label{fig:teaser}
\end{figure}

Compared to machine learning models, humans put themselves inside the 3D world and then perceive and interact with the surrounding environment from their ego-perspective (\Cref{fig:teaser}). Such situational awareness is a crucial difference between 2D and 3D visual understanding, and a key to achieving seamless understanding of \textit{spatial} concepts in more complex real-world environments. Several existing methods recognize the lack of positional understanding in 3D and propose new benchmarks and joint optimization functions~\cite{ma2022sqa3d}, or positional embedding methods~\cite{hong20233d} to enhance the overall reasoning performance. 

\begin{figure}
    \centering
    % \vspace{-2mm}
    \includegraphics[trim=210 125 210 115, clip=True, width=\linewidth]{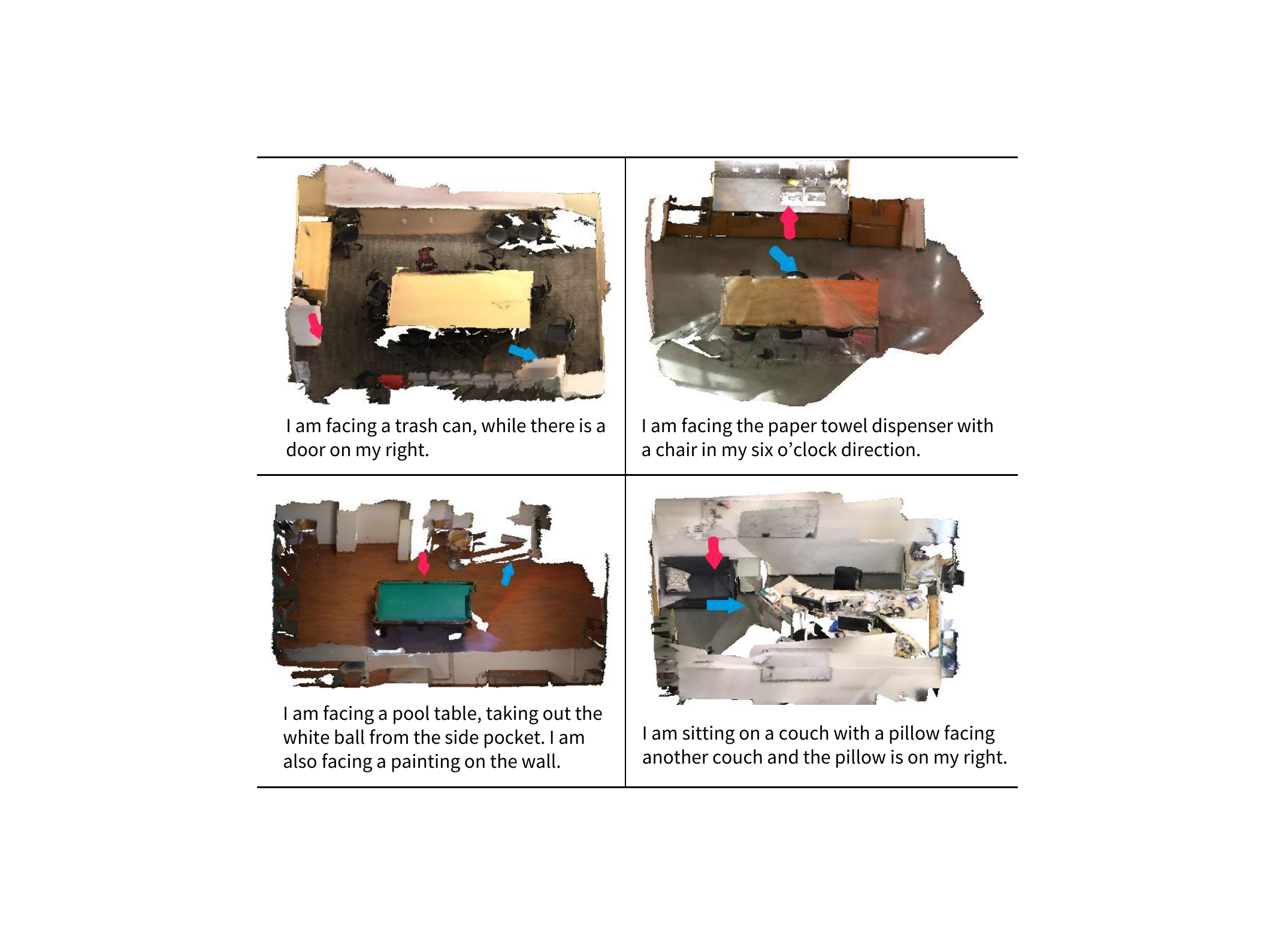}
    \vspace{-5mm}
    \caption{Situation estimation in existing methods~\cite{ma2022sqa3d} fails in most scenarios, indicating the missing registration between the situational descriptions and 3D embeddings.  \textcolor{Myred}{Red}: Ground truth (GT) vector. \textcolor{Myblue}{Blue}: Estimated vector.}
    \vspace{-4mm}
    \label{fig:inconsistency}
\end{figure}

However, the lack of an \emph{explicit} situation modeling and situation-grounded 3D reasoning method restricts them from obtaining a generalizable and consistent 3D vision-language (VL) representation. As shown in~\Cref{fig:inconsistency}, the situation prediction of the state-of-the-art method~\cite{ma2022sqa3d} (in blue) diverges significantly from the ground truth vectors (in red) in almost all scenes in the dataset~\cite{dai2017scannet}. Moreover, our pilot study in~\Cref{sec:pilot} also reveals that situational understanding, despite being very crucial in comprehending the context of questions, only plays a minor role in the final question answering (QA) performance of existing methods.

In this work, we propose SIG3D, a novel approach designed to \emph{precisely model and estimate an embodied agent's ego-location and orientation from a textual description}, before performing multi-modal QA tasks from the agent's ego-centric perspective, as shown in \Cref{fig:teaser}. Specifically, we leverage large-scale pretrained language and visual encoders to process the input text and 3D data, and fuse the tokens with attention modules to predict a situational vector. Previous attempts to directly predict the ego-situation are hindered by the expansive search space inherent in 3D environments. To address this challenge, we re-conceptualize the task as an \emph{anchor-based classification}, where visual tokens are regarded as anchor points, and a likelihood of position together with a set of rotation parameters are concurrently regressed for each visual token. After obtaining the situation estimation, we propose a situational alignment and a situation-guided tokens re-encoding strategy, to perceive the environment from the agent's intended perspective. These strategies enhance the visual tokens with more accurate situational awareness for subsequent QA tasks.

Experiments on two challenging 3D visual question answering (VQA) datasets~\cite{ma2022sqa3d,azuma2022scanqa} demonstrate the significant improvement in situation estimation and QA tasks of our model. In particular, we improve the accuracy of situation estimation by more than 30\%, and subsequent QA performance by up to 3\%. Further qualitative and quantitative analysis verifies our design choices and highlights the significance of situational awareness in 3D reasoning tasks. 

To sum up, our paper has the following contributions: (1) We recognize the lack of situational awareness as a significant oversight in existing research. To address this, we introduce \ourwork, a situation-grounded 3D VL reasoning architecture, specifically designed to fill this void.
(2) We propose an anchor-based approach to situation estimation, which effectively narrows the extensive search space in 3D environments for precise grounding of 3D positions and orientations with textual descriptions. Additionally, we investigate situational alignment and visual re-encoding mechanisms to leverage situational awareness for enhanced QA performance. (3) Our model demonstrates superior performance on two challenging datasets, SQA3D and ScanQA, surpassing the state of the art in both situation estimation and QA metrics. Ablation studies highlight the importance of situation-guided encoding, revealing its beneficial impact on general QA tasks.

\section{Related Work}
\label{sec:related_work}

\mypar{Vision Language Models (VLMs).} Early transformer-driven~\cite{vaswani2017attention} textual and visual encoders~\cite{kenton2019bert,dosovitskiy2020image} have facilitated great progress in recent vision language learning. Text-image contrastive models~\cite{radford2021learning,jia2021scaling} propose to align the feature space of two modalities with large-scale pretraining, fueling numerous downstream tasks from generalized open-vocabulary visual perception~\cite{kirillov2023segment,ghiasi2022scaling,li2022languagedriven} to text-to-image generation~\cite{rombach2022high}. Concurrently, some work uses text and vision encoders on separate modalities followed by feature fusion~\cite{dou2022empirical,kim2021vilt} for multi-modal reasoning tasks. Since the emergence of Large Language Models (LLMs)~\cite{brown2020language,touvron2023llama,zhang2022opt}, VLMs have experienced huge improvement with the help of LLMs as building blocks for multi-modal learning architectures. Specifically, recent work directly projects visual embeddings into language-space tokens as input to LLMs~\cite{lin2023towards,merullo2022linearly,zhang2023motiongpt}, or use the latent bottleneck structure for cross-modal visual decoding~\cite{alayrac2022flamingo,hong20233d,li2022blip,li2023blip}, or treat LLM layers as encoder blocks for various visual tasks~\cite{pang2023llm4vision}.

In the domain of visual question answering (VQA)~\cite{antol2015vqa,zhu2016visual7w}, recent work has pushed the frontier towards video understanding~\cite{lin2023towards,jia2022egotaskqa,jia2020lemma,wu2021star,datta2022episodic}, knowledge-based understanding~\cite{marino2019okvqa,schwenk2022okvqa,garderes2020conceptbert,gui2021kat,lin2022revive,wu2022multi}, and commonsense reasoning~\cite{zellers2019recognition}.
Despite the outstanding performance in 2D image interpretation, most existing methods lack the capability to generalize to 3D scenarios. In contrast, our work studies the representation of visual information and its fusion with language embeddings in the 3D domain by targeting on the 3D situation-guided visual language interpretation. 

\vspace{1.5mm}
\mypar{Grounding Language in 3D Space.} Compared with 2D images, knowledge such as spatial relationships, interactive exploration, and topological analysis -- which only exists in the 3D world -- provides additional challenges and opportunities to develop better language models with stronger commonsense reasoning capability grounded in the real-world 3D scenarios. In this direction, early work seeks to ground isolated objects~\cite{chen2019text2shape,achlioptas2019shapeglot} or objects within more complex scenes~\cite{chen2020scanrefer,achlioptas2020referit3d,huang2021text,feng2021free} using natural language descriptions. Recently, with more collected 3D vision language benchmarks, some work starts to explore language-guided 3D visual interpretation and reasoning on a diverse set of datasets, including 3D scene captioning~\cite{chen2021scan2cap}, open-vocabulary segmentation~\cite{peng2023openscene,ding2023pla,kerr2023lerf}, and question answering~\cite{azuma2022scanqa,hong20233dconcept,delitzas2023multiclip,ye20223d,zhu20233d,huang2024leo}. 

The success of LLMs also elicits the usage of them in 3D vision language reasoning for task decomposition~\cite{yang2023llm}, data generation, and multi-modal feature fusion~\cite{hong20233d}. Motivated by ScanQA~\cite{azuma2022scanqa}, SQA3D~\cite{ma2022sqa3d} takes the first step to explore the challenging 3D situational reasoning problem by developing a situated question answering benchmark, and proposing the first joint learning baseline on this benchmark. 
Our work highlights the uniqueness and significance of situational awareness in the 3D vision language learning paradigm, leading to notably better 3D situational grounding and question answering performance. 
\section{Pilot Study on Situational Reasoning}
\label{sec:pilot}

\begin{figure}
    \centering
    \vspace{-1mm}
    \includegraphics[trim=275 295 260 305, clip=True, width=\linewidth]{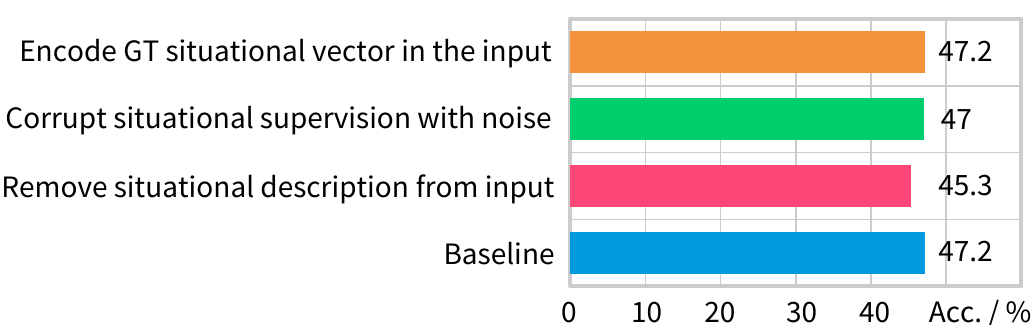}
    \vspace{-6mm}
    \caption{Results on variants of the representative SQA3D baseline method~\cite{ma2022sqa3d} demonstrate that situational understanding, despite being indispensable in perceiving the context of questions, makes negligible contribution in existing methods. This motivates a situation-guided 3D encoding mechanism in our model.}
    \vspace{-3mm}
    \label{fig:pilot}
\end{figure}

Despite highlighting the importance of situational understanding and reasoning, existing methods~\cite{ma2022sqa3d} fall short in providing effective situation estimation, as illustrated in~\Cref{fig:inconsistency}. This section delves into a pilot study that examines the impact of situational understanding on downstream reasoning tasks. The SQA3D baseline~\cite{ma2022sqa3d} incorporates situational descriptions and uses ground truth (GT) situational vectors for supervision in a direct regression task. We investigate three variants of this baseline to assess the effect of situational understanding. In the first variant, we \textit{remove the situational description and supervision} from the model, by passing in empty situational tokens. In another variant, we \textit{corrupt the situation supervision} by introducing very large Gaussian noise to the GT vectors to effectively randomize them. Finally, we try to \textit{encode the GT situational vector in the input} with learnable multi-layer perceptron (MLP) layers to form a GT situational token.

\Cref{fig:pilot} demonstrates the results of this study, revealing negligible changes in performance across these variants. Notably, corrupting the GT situational information or directly incorporating it results in only marginal alterations in the QA outcomes. Omitting the situational description entirely from the input results in a slight 2\% decrease in precision. However, in the absence of this information, the model resorts to random guessing when determining the correct answer, as all responses depend on the situation. The findings from~\Cref{fig:inconsistency,fig:pilot} collectively indicate a deficiency in existing methods regarding situation estimation and the application of situational understanding in subsequent reasoning tasks. These unresolved challenges motivate the development of our proposed method.

\section{Method}
\label{sec:method}

\begin{figure*}[!t]
    \centering
    \includegraphics[trim=99 205 90 210, clip=True, width=1\textwidth]{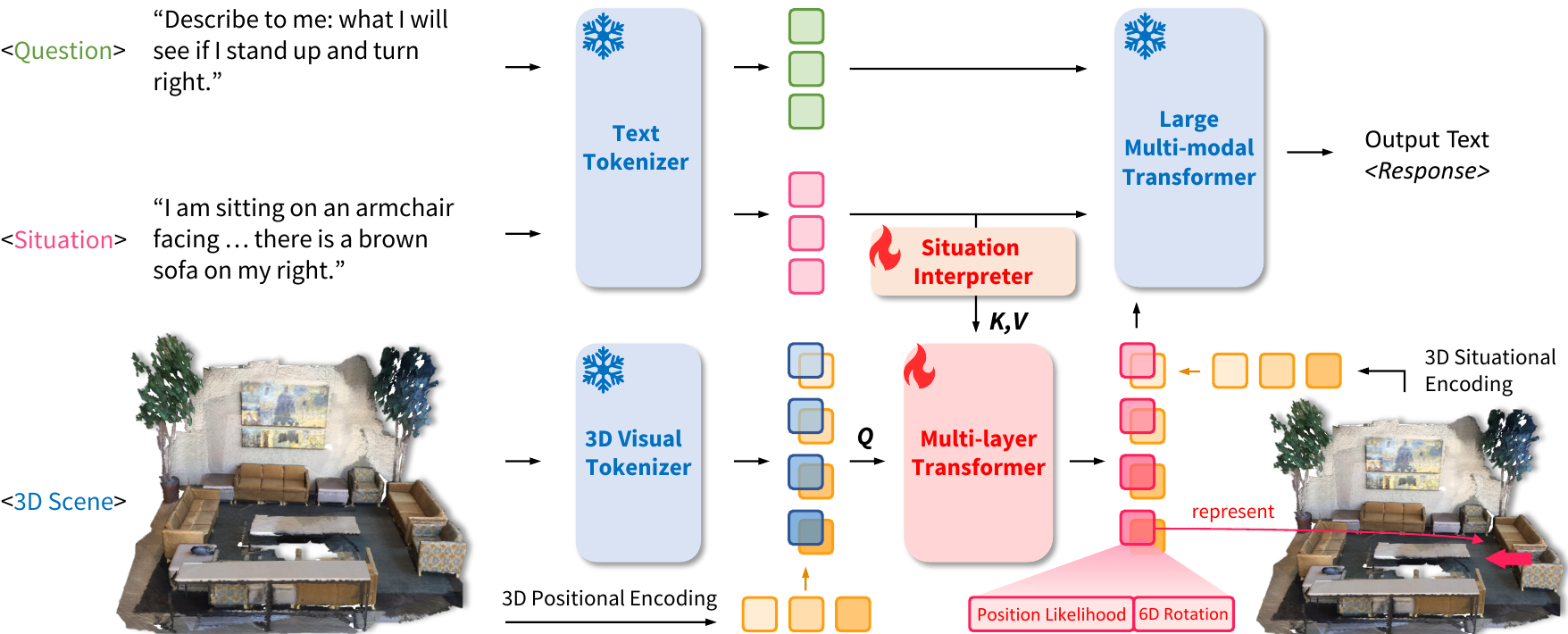}
    \vspace{-5mm}
    \caption{\textbf{Overview of our \ourwork model}, which includes 3D scene and text encoding, anchor-based situation estimation, situation-guided visual re-encoding, and multi-modal decoder modules. 
    We tokenize the 3D scene into voxels, treat each token as an anchor point, and query the text tokens to predict a token-level position likelihood and rotation matrix to locate the situational vector associated with the textual description. Then we update the scene tokens with situational position encoding (PE), and finally perform the 3D VL reasoning task with a large transformer decoder.
    }
    \vspace{-1mm}
    \label{fig:pipeline}
\end{figure*}

An overview of our approach \ourwork is illustrated in~\Cref{fig:pipeline}. Our method begins with a set of points that represent a 3D scene, accompanied by a situational description and a question that define the overall context of the problem. We tokenize them into separate token embeddings (\Cref{sec:method-token}), and ground the textual description in the 3D scene with a vector comprising location and orientation. We find direct single vector estimation to be challenging due to the vast and complex nature of the 3D search space, so we propose an anchor-based situation estimation strategy (\Cref{sec:method-situational-estimation}). Subsequently, we re-encode the visual tokens from the perspective of situational vectors, enhancing the situational awareness for downstream reasoning tasks (\Cref{sec:method-qa}). The finalized visual and textual tokens are fused by a transformer decoder to generate the final response.

\subsection{Visual and Textual Tokenization}\label{sec:method-token}

Leveraging input scene point clouds and textual prompts, our objective is to generate three distinct types of tokens: 3D visual tokens $z^{\mathrm{3D}}\in \mathbb{R}^{N_v \times C_v}$, situational tokens $z^{\mathrm{S}}\in \mathbb{R}^{N_s \times C_s}$, and question tokens $z^{\mathrm{Q}}\in \mathbb{R}^{N_q \times C_q}$. Each type of token is characterized by two primary components: $N$, representing the number of tokens, and $C$, encapsulating the feature embeddings. To tokenize and capture feature embeddings for situational input and questions, we employ a shared text tokenizer $\mathrm E^{\mathrm{TXT}}$ following prior methods~\cite{azuma2022scanqa,ma2022sqa3d}. We assume that situation and question prompts are separated in the input data. If not, LLMs~\cite{brown2020language} can be used to parse the textual input without changing the semantic meaning of the sentences. However, there is a lack of consensus on a standard 3D visual tokenization method $\mathrm E^{\mathrm{3D}}$ that is appropriate for the 3D VL reasoning task, prompting a more detailed exploration in the following paragraphs. 

\vspace{1.5mm}
\mypar{Visual Tokenization.} Given an input point cloud $\mathbf{p} \in \mathbb{R}^{N \times 3}$, most prior methods~\cite{azuma2022scanqa,ma2022sqa3d,delitzas2023multiclip} adopt a VoteNet~\cite{qi2019votenet} detector to acquire object-level tokens $z^{\mathrm{3D}}\in \mathbb{R}^{N_{\mathrm{obj}} \times C_{\mathrm{obj}}}$ as the visual representation, where $N_{\mathrm{obj}}$ is the number of object proposals, and $C_{\mathrm{obj}}$ is the object-level feature embeddings. However, we point out several problems with this abstraction strategy: (1) A detection-based tokenization method tends to ignore the non-object regions in the scene, which can be indispensable in some reasoning scenarios (\eg, carpets on the ground, ceiling, walls).  (2) After object-level abstraction, the visual representation losses the high-level information of the scene (\eg, the shape of the living room, the corner of the kitchen). (3) A supervised detector trained from scratch can only recognize objects within the training set (\eg, only 20 categories for ScanNet~\cite{dai2017scannet}), meaning that the method does not have zero-shot capability to reason about novel unseen objects that are inevitably common in real-world scenarios. 

In light of these, we adopt a pretrained open-vocabulary voxel-based tokenization method from OpenScene~\cite{peng2023openscene}. The scene is first discretized into regular small 3D voxels and fed into a visual encoder for feature extraction:
{%\small
  \setlength{\abovedisplayskip}{4pt}
  \setlength{\belowdisplayskip}{4pt}
  \setlength{\abovedisplayshortskip}{0pt}
  \setlength{\belowdisplayshortskip}{3pt}
\begin{align}\label{eq:feature_voxel}
    z^{\mathrm{3D}} = \mathrm E^{\mathrm{3D}}(\mathcal{V}(\mathbf{p})),
\end{align}}%
where $\mathcal{V}$ represents the voxelization process, and $\mathrm E^{\mathrm{3D}}$ is a Minkowski sparse 3D convolutional network~\cite{choy2019mink}. The sparse network is pretrained by distillation from CLIP~\cite{radford2021learning} embeddings of rendered multi-view 2D images, resulting in a feature map with better language alignment and 3D awareness. We take the upsampled bottleneck-layer feature embeddings from the encoder network, and compute the mean average over the $z$-axis (vertical) to project the voxels onto the $x$-$y$ plane and treat the feature grids in the resulting 2D feature map as our $N_v$ visual tokens. We find that this bird's-eye-view projection results in a more compact representation and improves the final performance.

\subsection{Situation Estimation}\label{sec:method-situational-estimation}

Given 3D visual tokens $z^{\mathrm{3D}}$ and situational tokens $z^{\mathrm{S}}$, our objective is to estimate the situational vector $\Vec{s}$ referred to by the situational description, which comprises a position component $s^{\mathrm{pos}}$ represented by coordinates $(x,y,z)$, and a rotation component $s^{\mathrm{rot}}$ represented by Euler angles ${(\theta,\psi,\phi)}$, where pitch angles $\psi$ are always defined as 0, meaning that situational vectors are defined to be parallel with the ground plane. The prior method~\cite{ma2022sqa3d} utilizes a transformer block to calculate the cross-attention feature between visual and language tokens, and directly regress a final situational vector from the averaged attention map. We find such a strategy producing very inaccurate estimates, as shown in~\Cref{fig:inconsistency}, due to the large search space in the entire 3D volume. Inspired by recent 3D object detection methods~\cite{yin2021center,zhou2022crossview,man2023bev}, we reduce the search space by \textit{turning the localization problem into a classification problem}. 

\vspace{1.5mm}
\mypar{Positional Embedding and Feature Fusion.} After the voxelization and 3D encoding process, each 3D token associates with a 3D position $(x,y,z)$ representing the center of its voxel. We first provide positional information to the model by generating learnable positional embeddings (PE) using a two-layer perceptron for each of the $N_v$ visual tokens, and add learnable positional embeddings to the token features $z^{\mathrm{3D}}$. We use a situation interpreter~\cite{reimers-2019-sentence-bert} to extract situational information, and ask the updated visual tokens to attend to these situational tokens with several transformer layers to produce the joint feature embeddings. 

\vspace{1.5mm}
\mypar{Anchor-based Situation Estimation.} We treat each output token of the feature fusion module as an anchor point, and use it to predict a position likelihood $p \in [0, 1]$ and a rotation estimation. Since each token has an associated 3D position $(x,y,z)$, the position likelihood $p$ indicates how likely the situational vector locates at the center of this token (voxel). We define a soft ground truth for this classification task with a Gaussian kernel, meaning that the closer a token is to the actual situational vector $s^{\mathrm{pos}}$, a higher ground truth probability $p$ will be assigned to that token. In order to counteract the sparse supervisory signal and increase the positive supervision around the vector position, we adopt the peak enlarging technique in CenterPoint~\cite{yin2021center}, where the size of the Gaussian kernel is increased (meaning that the $\sigma$ is increased) to allow denser supervision around the vector position. Furthermore, we explore different rotation representation and find that compared with quaternion and $(\sin\theta, \cos\theta)$ representations, the 6D vector proposed by~\cite{zhou20196d} achieves the best performance. Hence, we adopt a situation estimation head with MLP layers to output 7-dimensional vector for each of the tokens, where the first channel represents the position likelihood and the other six channels represent the 6D rotation matrix. We take the center of the token with the peak position likelihood as our estimated $s^{\mathrm{pos}}$, and convert its corresponding 6D rotation vector as our estimated $s^{\mathrm{rot}}$. The estimation can be equivalently represented as a rotation matrix $R$ and a translation matrix $T$. More discussion about the architecture and design choices is in~\Cref{sec:analysis_ablation}.

\subsection{Situation-guided Visual Encoding}\label{sec:method-qa}

After obtaining the situation estimation, we investigate a better approach to enhancing the generation of downstream responses, inspired by human cognitive processes. Intuitively, humans typically comprehend their immediate 3D environment by first interpreting their own situation in space, and then discerning their surroundings from an appropriate viewpoint. Our model is designed to emulate this natural strategy. Using the situational vector $\Vec{s}$, we adjust the coordinate system by repositioning the origin at $s^{\mathrm{pos}}$, and reorienting the axes according to $s^{\mathrm{rot}}$ so that the new y-axis is aligned with the indicated direction. We keep the $z$-axis vertically oriented and project the situational vectors onto the $x$-$y$ plane. This is in line with the format of the dataset~\cite{ma2022sqa3d}, where situational vectors are assumed to be parallel with the ground plane. Subsequently, we compute a new situation-guided PE for each of the $N_v$ visual tokens, similar to the learnable 3D PE outlined in~\Cref{sec:method-situational-estimation}. They allow the model to grasp the positional interrelations from the perspective of the current situation. These situational embeddings are added to the output embeddings of the situation estimation module, which consists of blocks featuring self-attention layers for visual tokens, succeeded by cross-attention layers that bridge visual and situational information. This structure allows for the re-encoding of visual tokens under the influence of situation and question context, guiding the model to assign higher weights to situation-related and question-related visual tokens. The output, termed \textit{situation-guided visual tokens}, embodies this re-contextualized understanding.

\subsection{Question Answering Head}

We follow existing methods~\cite{hong20233d} to use a large vision-language decoder to fuse the final visual and textual tokens and generate textual response to the input question. We explore both auto-regressive response generation and classification-based answer prediction~\cite{ma2022sqa3d,azuma2022scanqa}. For classification, we predict a vector $v^{\mathrm{ans}} \in \mathbb{R}^{n_a}$ for the candidates of $n_a$ answers in the training set following~\cite{azuma2022scanqa}.

\begin{table*}
\begin{subtable}[h]{\linewidth}
\centering
\resizebox{0.7\linewidth}{!}{
\begin{tabular}{lc@{\hspace{8mm}}c@{\hspace{6mm}}c@{\hspace{6mm}}c@{\hspace{6mm}}c@{\hspace{6mm}}c@{\hspace{6mm}}c}
\toprule
\multirow{2}{*}[-1mm]{Model} & \multicolumn{6}{c}{Question Breakdown} & \multirow{2}{*}[-1mm]{Overall}\\
\cmidrule(r{2mm}){2-7}
& What & Is & How & Can & Which & Other & \\
\midrule
GPT-3~\cite{brown2020language}          & \textbf{39.7} & 46.0 & 40.5 & 45.6 & 36.1 & 38.4 & 41.0\\
ClipBERT~\cite{lei2021less}             & 30.2 & 60.1 & 38.7 & 63.3 & 42.5 & 42.7 & 43.3\\
MCAN~\cite{yu2019deep}                  & 28.9 & 59.7 & 44.1 & 68.3 & 40.7 & 40.5 & 43.4\\
ScanQA~\cite{azuma2022scanqa}           & 28.6 & 65.0 & 47.3 & 66.3 & 43.9 & 42.9 & 45.3\\
SQA3D~\cite{ma2022sqa3d}                & 33.5 & 66.1 & 42.4 & 69.5 & 43.0 & 46.4 & 47.2\\
Multi-CLIP~\cite{delitzas2023multiclip} & -    & -    & -    & -    & -    & -    & 48.0\\
LM4Vision~\cite{pang2023llm4vision}     & 34.3 & 67.1 & 48.2 & 68.3 & {48.9} & 45.6 & 48.1\\
3D-LLM~\cite{hong20233d}                & 36.5 & 65.6 & 47.2 & 68.8 & 48.0 & 46.3 & 48.1\\
3D-VisTA~\cite{zhu20233d}               & 34.8 & 63.3 & 45.4 & 69.8 & {47.2} & \textbf{48.1} & 48.5\\
\midrule
 % SIG3D (Ours)                           & 35.6 & \textbf{67.2} & \textbf{48.5} & \textbf{71.4} & \textbf{49.1} & 45.8 & \textbf{50.9}\\
 SIG3D (Ours)                           & 35.6 & \textbf{67.2} & \textbf{48.5} & \textbf{71.4} & \textbf{49.1} & 45.8 & \textbf{52.6}\\
\bottomrule
\end{tabular}}
\end{subtable}
\vspace{-2mm}
\caption{Our proposed SIG3D achieves state-of-the-art performance on the SQA3D benchmark~\cite{ma2022sqa3d}. We perform the best on ``Is,'' ``How,'' and ``Can'' breakdown types of questions, as well as the average accuracy with EM@1 metric. The results are reported on the test set.}
\label{tab:sqa3d_em}
\vspace{-3mm}
\end{table*}
\begin{table}
\begin{subtable}[h]{\linewidth}
\centering
\resizebox{\linewidth}{!}{
\begin{tabular}{l@{\hspace{2mm}}c@{\hspace{2mm}}c@{\hspace{3mm}}c@{\hspace{2mm}}c}
\toprule
\multirow{2}{*}[-1mm]{Model} & \multicolumn{2}{c}{Localization}&\multicolumn{2}{c}{Orientation}\\
\cmidrule(r{2.5mm}){2-3}\cmidrule(l{-0.5mm}){4-5}
& Acc@0.5m& Acc@1.0m& Acc@15° & Acc@30°\\
\midrule
Random       & 7.2 & 25.8 & 8.4 & 16.9 \\
SQA3D~\cite{ma2022sqa3d}        & 9.5 & 29.6 & 8.7 & 16.5  \\
SQA3D (\textit{separate})       & 10.3 & 31.4 & 17.1 & 22.8  \\
3D-VisTA~\cite{zhu20233d}       & 11.7 & 34.5 & 16.9 & 24.2  \\
\midrule
SIG3D (Ours)  & \textbf{27.4} & \textbf{59.1} & \textbf{28.7} & \textbf{42.5} \\
\bottomrule
\end{tabular}}
\end{subtable}
\vspace{-2mm}
\caption{Our proposed method SIG3D performs significantly better than prior methods~\cite{ma2022sqa3d} in the situation estimation task. ``Acc@0.5m'' stands for localization accuracy with 0.5m threshold. ``Acc@15°'' represents orientation accuracy with 15° threshold. \textit{separate} means disabling other tasks to let the model focus on situation estimation only.}
\label{tab:sqa3d_situation}
\vspace{-2mm}
\end{table}

\section{Analysis in 3D VQA Task}
\label{sec:analysis}

We evaluated \ourwork for 3D VL reasoning on two challenging benchmarks, addressing both visually-oriented situation estimation and textual-focused QA tasks. We present a detailed examination of the implementation strategies adopted, the datasets employed, and the metrics applied in our research. For exhaustive understanding, implementation, training details, and other additional information are available in the \textbf{supplementary material}.

\vspace{1.5mm}
\mypar{Datasets.} We evaluate our method on SQA3D~\cite{ma2022sqa3d} and ScanQA~\cite{azuma2022scanqa}, two challenging indoor 3D VQA datasets. Both datasets are derived from the ScanNet dataset~\cite{dai2017scannet}, serving as the foundational source for their 3D scenes. SQA3D features over 33K question-answer pairs for the 3D VQA task and 26K unique situational descriptions for the situation estimation task. Each entry in this dataset includes a 3D scene point cloud, a situational description, a question, and pertinent annotations. ScanQA consists of over 41K question-answer pairs, without situational descriptions and situational annotations. We use it to demonstrate the generalizability of our method on general QA tasks. We use the splits provided by these datasets.

\vspace{1.5mm}
\mypar{Evaluation Metrics.} For SQA3D, in order to compare with baseline methods~\cite{ma2022sqa3d,pang2023llm4vision,zhu20233d}, we use a shallow transformer decoder task head to perform the answer classification task, and evaluate the performance with exact matches (EM@1), which is equivalent to Top-1 answer accuracy. We also provide EM@1 on a breakdown of question types, including ``What,'' ``Is,'' ``How,'' ``Can,'' ``Which,'' and ``Other,'' based on the first word in the question sentence. Additionally, we evaluate situation estimation performance with localization accuracy and orientation accuracy. In both tasks, we use accuracy within different distance or angle thresholds as our metrics. For example, ``Acc@0.5m'' means accuracy of location estimation when positive threshold is set to 0.5 meter. For ScanQA, we perform auto-regressive answer generation with large transformer decoder~\cite{hong20233d}, and evaluate with BLEU~\cite{papineni2002bleu}, ROUGE~\cite{lin2004rouge}, METEOR~\cite{banerjee2005meteor}, and CIDEr~\cite{vedantam2015cider} metrics.

% \begin{table*}
% \begin{subtable}[h]{\linewidth}
% \centering
% \resizebox{0.75\linewidth}{!}{
% \begin{tabular}{lc@{\hspace{6mm}}c@{\hspace{6mm}}c@{\hspace{6mm}}c@{\hspace{6mm}}c@{\hspace{6mm}}c}
% \toprule
% Model & BLEU-1& BLEU-4& ROUGE& METEOR& CIDEr &EM@1
% \\
% \midrule
% BLIP2-MultiView~\cite{li2023blip}               & 29.7 & 5.9  & 26.6 & 11.3 & 45.7 &13.6 
% \\
% flamingo-SingleImage~\cite{alayrac2022flamingo} & 23.8 & 8.5  & 29.6 & 10.7 & 52.0 &16.9 
% \\
% Flamingo-MultiView~\cite{alayrac2022flamingo}   & 25.6 & 8.4  & 31.1 & 11.3 & 55.0 &18.8 
% \\
% VoteNet+MCAN~\cite{yu2019deep}                  & 28.0 & 6.2  & 29.8 & 11.4 & 54.7 &17.3 
% \\
% ScanRefer+MCAN~\cite{yu2019deep}                & 26.9 & 7.9  & 30.0 & 11.5 & 55.4 &18.6 
% \\
% ScanQA~\cite{azuma2022scanqa}                   & 30.2 & 10.1 & 33.3 & 13.1 & 64.9 &21.0 
% \\
% 3D-LLM~\cite{hong20233d}                        & \textbf{39.3} & 12.0 & 35.7 & \textbf{14.5} & \textbf{69.4} &20.5 
% \\
% \midrule
% SIG3D                   & 36.9 & \textbf{12.4} & \textbf{35.9} & 13.4 & 68.8  &\textbf{22.3} \\
% \bottomrule
% \end{tabular}}
% \end{subtable}
% \vspace{-1.5mm}
% \caption{Performance of \ourwork on ScanQA dataset~\cite{azuma2022scanqa} is on-par with the state-of-the-art with large-scale text-3D pre-training. 3D-LLM~\cite{zhu20233d} leverages pretrained 2DVL foundation models and LLM models~\cite{li2023blip,alayrac2022flamingo,brown2020language,li2022blip}, and is pretrained on a large-scale held-in 3D-text dataset before the finetuning on ScanQA.}
% \label{tab:scanqa}
% \vspace{-3mm}  
% \end{table*}

\begin{table}
\begin{subtable}[h]{\linewidth}
\centering
\resizebox{\linewidth}{!}{
\begin{tabular}{lc@{\hspace{3mm}}c@{\hspace{3mm}}c@{\hspace{3mm}}c@{\hspace{3mm}}c}
\toprule
Model & BLEU-1 & BLEU-4& ROUGE& METEOR& CIDEr\\
\midrule
BLIP2~\cite{li2023blip}               & 29.7 & 5.9  & 26.6 & 11.3 & 45.7 
\\
Flamingo~\cite{alayrac2022flamingo}   & 25.6 & 8.4  & 31.1 & 11.3 & 55.0
\\
VN+MCAN~\cite{yu2019deep}             & 28.0 & 6.2  & 29.8 & 11.4 & 54.7 
\\
SR+MCAN~\cite{yu2019deep}             & 26.9 & 7.9  & 30.0 & 11.5 & 55.4
\\
ScanQA~\cite{azuma2022scanqa}         & 30.2 & 10.1 & 33.3 & 13.1 & 64.9 
\\
3D-LLM~\cite{hong20233d}              & 39.3 & 12.0 & 35.7 & \textbf{14.5} & \textbf{69.4}
\\
\midrule
SIG3D                                 & \textbf{39.5} & \textbf{12.4} & \textbf{35.9} & 13.4 & 68.8 \\
\bottomrule
\end{tabular}}
\end{subtable}
\vspace{-1.5mm}
\caption{Performance of \ourwork on the ScanQA dataset~\cite{azuma2022scanqa} is on-par with the state of the art with large-scale text-3D pretraining. VN and SR stand for VoteNet and ScanRefer, respectively. 3D-LLM~\cite{zhu20233d} leverages pretrained 2D VL foundation models and LLM models~\cite{li2023blip,alayrac2022flamingo,brown2020language,li2022blip}, and is pretrained on a large-scale held-in 3D-text dataset before the finetuning on ScanQA.}
\label{tab:scanqa}
\vspace{-2mm}  
\end{table}

\subsection{Situated Question Answering}

\mypar{Baselines.} Our study involves a comparative analysis with a range of representative baselines on the SQA3D dataset. In particular, we evaluate against \textbf{GPT-3}~\cite{brown2020language}, \textbf{ClipBERT}~\cite{lei2021less}, and \textbf{MCAN}~\cite{yu2019deep}, which are, as reported in prior work~\cite{ma2022sqa3d}, baselines focused on language-only, 2D video, and 2D image QA, respectively. For GPT-3, we follow SQA3D~\cite{ma2022sqa3d} to convert the visual input into a caption using Scan2Cap~\cite{chen2021scan2cap} for LLMs to process. \textbf{ScanQA}~\cite{azuma2022scanqa} represents a 3D QA baseline that ignores the situational input. Both \textbf{SQA3D}~\cite{ma2022sqa3d} and \textbf{Multi-CLIP}~\cite{delitzas2023multiclip} employ situational descriptions and annotations for direct regression tasks. \textbf{LM4Vision}~\cite{pang2023llm4vision} utilizes LLMs as visual and textual encoders. Additionally, \textbf{3D-VisTA}~\cite{zhu20233d} undergoes a pretraining procedure on their large-scale 3D scene-text dataset, ScanScribe, prior to the finetuning on this dataset.

\begin{table*}[t]
\centering
\vspace{2mm}
\begin{subtable}[t]{0.35\linewidth}
\centering
\caption{Number of Visual Tokens}
\vspace{-1pt}
\scalebox{0.9}{
\begin{tabular}{lccc}
    \toprule
    &Acc@1.0m & Acc@30° & EM@1\\\midrule
    128 & 48.9 & 38.2 & 49.2 \\ 
    \rowcolor{LightGrey}
    256 & \textbf{59.1} & \textbf{42.5} & \textbf{50.9} \\ 
    512 & 57.8 & 42.1 & 50.7 \\
    \bottomrule
\end{tabular}
}
\label{tab:ablation_architecture_visual_token}
\end{subtable}
~
\begin{subtable}[t]{0.3\linewidth}
\centering
\caption{Voxel size (\textit{in meters})}
\vspace{-1pt}
\scalebox{0.9}{
\begin{tabular}{lcc}
    \toprule
    &Acc@1.0m &EM@1\\\midrule
    0.01  & 54.1 & 49.5 \\ 
    \rowcolor{LightGrey}
    0.02  & \textbf{59.1} & \textbf{50.9} \\
    0.05 & 47.3 & 48.8 \\ 
    \bottomrule
\end{tabular}
}
\label{tab:ablation_architecture_voxel_size}
\end{subtable}
~
\begin{subtable}[t]{0.3\linewidth}
\centering
\addtolength{\tabcolsep}{1pt} 
\caption{Rotation representation}
\vspace{-1pt}
\scalebox{0.9}{
\begin{tabular}{lcc}
    \toprule
    & Acc@30° & EM@1\\\midrule
    Quaternion & 31.4 & 50.0 \\ 
    \rowcolor{LightGrey}
    6D vector & 42.5 & \textbf{50.9} \\
    $\sin\theta, \cos\theta$ & \textbf{42.6} & 50.6 \\
    \bottomrule
\end{tabular}
}
\label{tab:ablation_architecture_rotation}
\end{subtable}
\caption{Ablation study validates that our various design choices improve the performance. ``Acc@1.0m,'' ``Acc@30°,'' and ``EM@1'' are accuracy (\%) for localization estimation, orientation estimation, and QA tasks, respectively. Our settings are marked in \colorbox{LightGrey}{gray}.}
\vspace{-3mm}
\label{tab:ablation_architecture}
\end{table*}

\begin{table}
\begin{subtable}[h]{\linewidth}
\centering
\resizebox{\linewidth}{!}{
\begin{tabular}{l|ccc}
    \toprule
    &Acc@1.0m & Acc@30° & EM@1\\\hline
    \rowcolor{LightGrey}\multicolumn{4}{l}{\textit{3D Vision Encoder}} \\
    % \rowcolor{LightGrey}
    Text-only (\textit{no vision input}) &- & - & 47.5 \\ 
    VoteNet~\cite{qi2019votenet} & 37.4 & 28.2 & 49.1 \\ 
    3DETR~\cite{misra2021end} & 47.2 & 29.1 & 49.4 \\ 
    OpenScene - OpenSeg~\cite{peng2023openscene}  & 57.5 & 41.6 & 50.2 \\ 
    OpenScene - LSeg~\cite{peng2023openscene}  & \textbf{59.1} & \textbf{42.5} & \textbf{50.9}\\\hline
    \rowcolor{LightGrey}\multicolumn{4}{l}{\textit{Language Tokenizer / Encoder}} \\
    GloVe + LSTM~\cite{hochreiter1997lstm,pennington2014glove} & 44.3 & 30.9 & 48.7 \\ 
    SBERT - MiniLM~\cite{reimers-2019-sentence-bert}  & 56.1 & 38.6 & 49.4 \\ 
    SBERT - MPNet~\cite{reimers-2019-sentence-bert} & 55.9 & 40.6 & 49.7 \\ 
    SBERT - MPNet (\textit{finetune})  & \textbf{59.1} & \textbf{42.5} & \textbf{50.9}\\
    \bottomrule
\end{tabular}}
\end{subtable}
\vspace{-2mm}
\caption{Performance of SIG3D improves with stronger visual and language encoders. We find that the open-vocabulary point encoder and MPNet-based sentence BERT (SBERT) leads to the best performance. ``Acc@1.0m'' and ``Acc@30°'' stand for localization and orientation accuracy in the situation estimation task, respectively. ``EM@1'' demonstrates the exact match metric in the QA task.}
\label{tab:ablation_encoder}
\vspace{-2mm}
\end{table}

\vspace{1.5mm} 
\mypar{Situation Estimation.} As shown in \Cref{tab:sqa3d_situation}, our work performs significantly better than the state of the art~\cite{ma2022sqa3d,zhu20233d} in both localization and orientation estimation tasks. For 3D-VisTA~\cite{zhu20233d}, we use a pretrained model and finetune a new situation head with the SQA3D dataset following~\cite{ma2022sqa3d}. We also report a \textit{random} baseline, in which we randomly sample position and orientation from a uniform distribution as a lower-bound performance. Note that the original SQA3D performs only marginally better than the random baseline, meaning that it does not acquire any situational awareness, despite having the situation estimation loss. Disabling the QA task and asking the model to exclusively focus on the situation estimation task results in a slight better performance. Our method, with the anchor-based position likelihood estimation, results in much better understanding of the 3D situational relationship. Our method also outperforms 3D-VisTA, which is pretrained on a large-scale 3D-text dataset, indicating that large pretraining alone is not enough to address the situational awareness problem. Note that we do not include the random baseline performance reported in ~\cite{ma2022sqa3d}, because each value is obtained by generating three random values and taking the \textit{closest} one to the ground true, and thus it does not reflect a true ``random'' baseline.

\vspace{1.5mm}  
\mypar{Situated Question Answering.} \ourwork outperforms prior methods in most question breakdown categories and overall accuracy, as shown in \Cref{tab:sqa3d_em}. Our work achieves leading results \textit{without large-scale pretraining} (compared with 3D-VisTA) and LLMs (compared with GPT-3), indicating its superiority in situational awareness. Note that the LLM baseline GPT-3 achieves the best performance on the ``What'' category, suggesting the potential of a stronger language encoder in interpreting complicated questions.

\subsection{General Question Answering on ScanQA} 

\mypar{Baselines.} We compare with 2D image VQA \textbf{MCAN}-based baselines~\cite{yu2019deep}, \textbf{ScanQA}~\cite{azuma2022scanqa}, \textbf{3D-LLM}~\cite{hong20233d} which leverages large-scale pretrained 2D VLMs and LLMs as backbone models, and \textbf{3D-VisTA}~\cite{zhu20233d} pretrained on their proposed large-scale 3D-text dataset. 

\vspace{1.5mm} 
\mypar{Question Answering.} As shown in~\Cref{tab:scanqa}, despite that the questions do not explicitly require situational understanding to answer in ScanQA, \ourwork achieves comparable results with state-of-the-art methods without the large-scale 3D-text pretraining and powerful 2D VLM and LLM backbone models. Our work pretrained on SQA3D~\cite{ma2022sqa3d} leads to higher performance on BLEU-1, BLEU-4, and ROUGE metrics, showing its generalizability on general 3D QA scenarios.

\begin{figure*}[!t]
    \centering
    \includegraphics[trim=5 115 10 87, clip=True, width=1\textwidth]{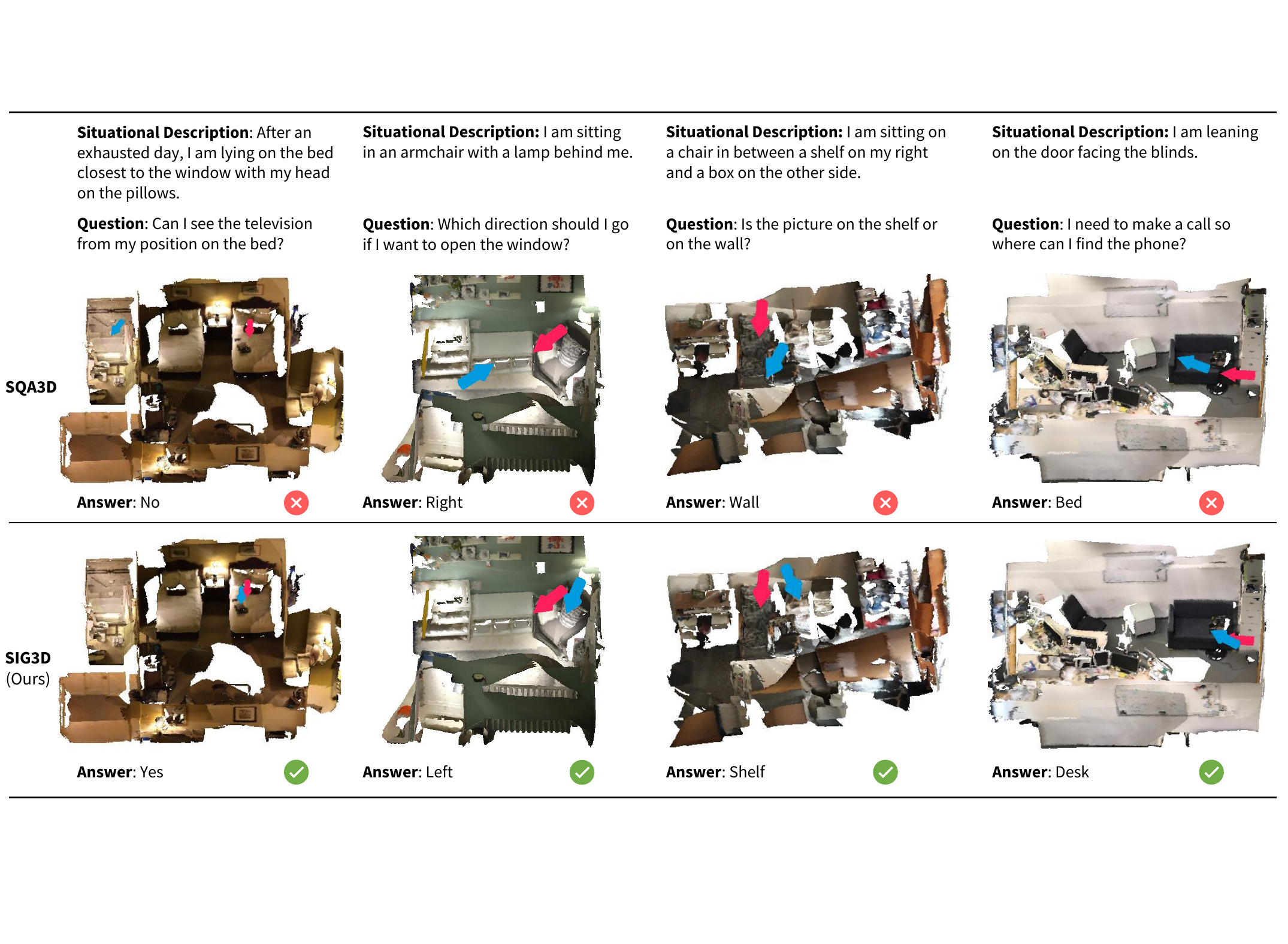}
    \vspace{-6mm}
    \caption{Qualitative results demonstrate significant improvement of \ourwork over prior methods. The first row is results from SQA3D~\cite{ma2022sqa3d}, and the second row is results from our method. In the 3D scene, \textcolor{Myred}{red}: Ground truth (GT) vector, and \textcolor{Myblue}{blue}: Estimated vector.
    }
    \vspace{-3mm}
    \label{fig:qualitative}
\end{figure*}

\subsection{Ablation Study and Analysis}\label{sec:analysis_ablation}

\mypar{Vision and Language Encoders.} We study the impact of different visual and textual tokenizers in~\Cref{tab:ablation_encoder}. It is observed that the open-vocabulary visual encoder (OpenScene) outperforms detection-based encoders (such as VoteNet and 3DETR) across all metrics. This superior performance of OpenScene is attributed to the limitations of 3D detectors, which are typically trained on a limited set of object categories, rendering them less effective in recognizing novel objects mentioned in textual prompts. Regarding language encoders, our findings indicate that a stronger backbone correlates with better performance, primarily due to its improved capability to interpret complex textual inputs. This leads to the suggestion of integrating LLMs with our method to potentially further enhance performance, an avenue we intend to explore in future research.

\begin{table}
\begin{subtable}[h]{\linewidth}
\centering
\resizebox{\linewidth}{!}{
\begin{tabular}{l|ccc}
    \toprule
    &Acc@1.0m & Acc@30° & EM@1\\\hline
    Baseline (\textit{joint optimization}) & 29.5 & 23.1 & 47.7 \\\hline
    \rowcolor{LightGrey}\multicolumn{4}{l}{\textit{How to achieve better situation estimation}} \\
    + 3D PE        & 38.8 & 23.6 & 47.8 \\
    + 6D Representation & 38.5 & 27.4 & 47.7\\
    + Anchor-based Estimation       & 58.8 & 41.9 & 48.2\\\hline
    \rowcolor{LightGrey}\multicolumn{4}{l}{\textit{How to utilize situation estimation for better QA}} \\
    + 3D Situational PE          & 58.9 & 41.8 & 50.0 \\ 
    + Visual Token Re-encoding      & \textbf{59.1} & \textbf{42.5} & \textbf{50.9} \\\hline
    \rowcolor{LightGrey}\multicolumn{4}{l}{\textit{Oracle Model (Ground Truth Situation Information)}} \\
    Situation as direct input            & 100 & 100 & 47.7 \\
    Situation as intermediate input      & 100 & 100 & 53.9 \\
    \bottomrule
\end{tabular}}
\end{subtable}
\vspace{-2mm}
\caption{Ablation study verifies that our proposed modules lead to better situation estimation and better QA performance.}
\label{tab:ablation_situation}
\vspace{-2mm}
\end{table}

% \begin{table*}
% \begin{subtable}[h]{0.59\linewidth}
% \centering
% \resizebox{\linewidth}{!}{
% \begin{tabular}{l|ccc}
%     \toprule
%     &Acc@1.0m & Acc@30° & EM@1\\\hline
%     Baseline (\textit{joint optimization}) &- & - &- \\\hline
%     \rowcolor{LightGrey}\multicolumn{4}{l}{\textit{How to achieve better situation estimation}} \\
%     + 3D Positional Encoding        & - & - & - \\
%     + Sine/Cosine Representation    & - & - & -\\
%     + Anchor-based Estimation       & - & - & -\\\hline
%     \rowcolor{LightGrey}\multicolumn{4}{l}{\textit{How to utilize situation estimation for better QA}} \\
%     + Two-stage Learning            &- & - &- \\ 
%     + Situational Encoding          &- & - &- \\ 
%     + Visual Token Reweighting      & \textbf{42.1} & \textbf{61.8} &47.5\\\hline
%     \rowcolor{LightGrey}\multicolumn{4}{l}{\textit{Oracle Models (Ground Truth Situation Information)}} \\
%     Situation as direct input       & 100  & 100 &  \\
%     Situation as 2-stage input      & 100  & 100 &  \\
%     \bottomrule
% \end{tabular}}
% \end{subtable}
% \begin{subfigure}[h]{0.355\linewidth}
% \caption{3D VQA}
% \resizebox{0.99\linewidth}{!}
% {
% \includegraphics[trim=380 270 380 270, clip=True, width=0.98\linewidth]{figures/increased_data.pdf}
% }
% \label{tab:3dqa}
% \end{subfigure}
% \vspace{-2mm}
% \caption{Our proposed method SIG3D XXX.}
% \label{tab:ablation_situation}
% \vspace{-3mm}
% \end{table*}

\vspace{1.5mm}
\mypar{Situational Awareness.} In \Cref{tab:ablation_situation} we verify the crucial role of situational awareness in the 3D VL task. Firstly, we show that 3D PE, 6D rotation estimation, and anchor-based position estimation all lead to much better position and orientation estimation performance. We further establish that situational PE and visual token re-encoding modules lead to better utilization of the predicted situational vector for the QA task. Additionally, We design two oracle models under the assumption of having access to the ground truth situational vector as input. The outcomes from these models reveal a critical insight: the model fails to effectively interpret situational information when it is directly incorporated into the input visual embeddings. This underlines the necessity of the intermediate representation and encoding mechanism we have proposed, affirming its importance in achieving optimal 3D VL task performance.

\vspace{1.5mm}
\mypar{Architectural Design.} We explore different architectural design choices of our model in \Cref{tab:ablation_architecture}. We find that the number of visual tokens sampled from the visual feature embeddings affects the performance of both situation estimation and QA tasks. Sampling fewer visual tokens increases the risk of missing the region of significance, while sampling more does not lead to a better performance as well. We study the size of voxels and find 0.02m to be the most effective choice, as the OpenScene~\cite{peng2023openscene} backbone is pretrained with the same voxel size. We also find that the  $(\sin\theta, \cos\theta)$ and 6D vector representations perform a lot better than quaternion in the rotation estimation task. This is consistent with the finding reported in~\cite{zhou20196d}.

\subsection{Qualitative Analysis}\label{sec:qualitative}

Finally, we demonstrate some qualitative results of our \ourwork in ~\Cref{fig:qualitative}. We show the ground truth and estimated situational vectors in red and blue, respectively, in their corresponding 3D scenes. We also print the answers with a red cross or green checkmark indicating the correctness. It is clear that our method performs significantly better in situation estimation tasks, resulting in vectors very close to the ground truth in both position and orientation perspectives. Better situational awareness also aids the complicated embodied navigation and commonsense QA activities. This further demonstrates great potential of our method in the development of indoor robotics and/or conversational agents.

\vspace{1.5mm}
\mypar{Supplementary Material.} The supplementary section offers an extensive analysis, encompassing a detailed examination of the \textit{3D visual token activation changes} pre- and post-situational re-encoding. Additionally, it includes a comprehensive collection of \textit{positive and negative samples}, an insightful \textit{failure case analysis}, and a forward-looking discussion on \textit{limitations and future work}.
\section{Conclusion}
\label{sec:conclusions}

In this paper, we introduce SIG3D, a situation-aware vision language model for 3D reasoning tasks. We propose to represent 3D scenes as feature tokens, treat tokens as anchor points to estimate a situational vector from a textual description, and use the estimated situation as guidance to align and re-encode the visual tokens to enhance the features for reasoning tasks. We observe consistent and significant performance gains on both situation estimation and question answering tasks. 

\noindent{\footnotesize\textbf{Acknowledgement.} This work was supported in part by NSF Grant 2106825, NIFA Award 2020-67021-32799, the Jump ARCHES endowment, and the IBM-Illinois Discovery Accelerator Institute. This work used NVIDIA GPUs at NCSA Delta through allocations CIS220014, CIS230012, and CIS230013 from the ACCESS program.}

{
    \small
    \bibliographystyle{ieeenat_fullname}
    \bibliography{main}
}

\clearpage
\setcounter{page}{1}
\maketitlesupplementary

\renewcommand{\thesection}{\Alph{section}}
\renewcommand{\thefigure}{\Alph{figure}}
\setcounter{section}{0}
\setcounter{figure}{0}

\section{Implementation Details} \label{sec:more_implementation_details}

Here we provide more details about our model. 

\vspace{1.5mm}
\mypar{Visual and Textual Encoders.} We use OpenScene~\cite{peng2023openscene} (the 3D distilled variant) as our visual encoder, which incorporates distilled CLIP features into a 3D Minkowski CNN backbone originally designed for the 3D semantic segmentation task. We use the default 0.02m voxel size to discretize a point cloud into 3D voxels and disable the scaling and elastic distortion augmentation methods during the voxelization process. The 3D architecture is the predefined \textit{MinkUNet18A}~\cite{choy2019mink}. The number of visual tokens $N_v$ is 256. Additionally, we use the Sentence-BERT~\cite{reimers-2019-sentence-bert} MPNet variant as our text tokenizer and encoder. We use the fixed batch padding strategy and set the length to be 100 for both situational tokens $N_s$ and question tokens $N_q$. The feature embedding sizes for all three types of tokens are set to 768. The 256 dimensional output of the OpenScene backbone is projected to the 768 hidden size with a 1x1 Conv layer. We freeze the OpenScene backbone, and finetune only the last layer of the textual backbone during our training process.

\vspace{1.5mm}
\mypar{Fusion and Decoder Models.} We use a 4-layer MCAN Transformer~\cite{yu2019deep,vaswani2017attention} as the fusion block of the visual and situational tokens. The learnable positional embeddings and situational embeddings are composed of a 2-layer MLP: first from dimension 3 to 128, and then from 128 to the target dimension. We use BLIP-2~\cite{li2023blip} as the large multi-modal tranformer for final response generation, simiar to 3D-LLM~\cite{hong20233d}.

\vspace{1.5mm}
\mypar{Training Details.} We train our model with the AdamW~\cite{loshchilov2017decoupled} optimizer with $\beta_1=0.9$, $\beta_2=0.999$, and $\epsilon=1\mathrm{e}$-8. We use a batch size of 16 and set the initial learning rate to be 2$\mathrm{e}$-5. The weight decay is set to 0.05, and we disable the weight decay on the layernorm layers and all bias parameters following~\cite{lin2023towards}. We decrease the learning rate by 10 times after the 10-th and the 20-th epochs. We train the model for a total of 50 epochs on a single NVIDIA A100 GPU.

\section{Enhanced Vision Token Activation Through Situational Re-encoding}\label{sec:supp_activation}    

\begin{figure*}[t]
    \centering
    \includegraphics[trim=5 550 60 40, clip=True, width=0.95\linewidth]{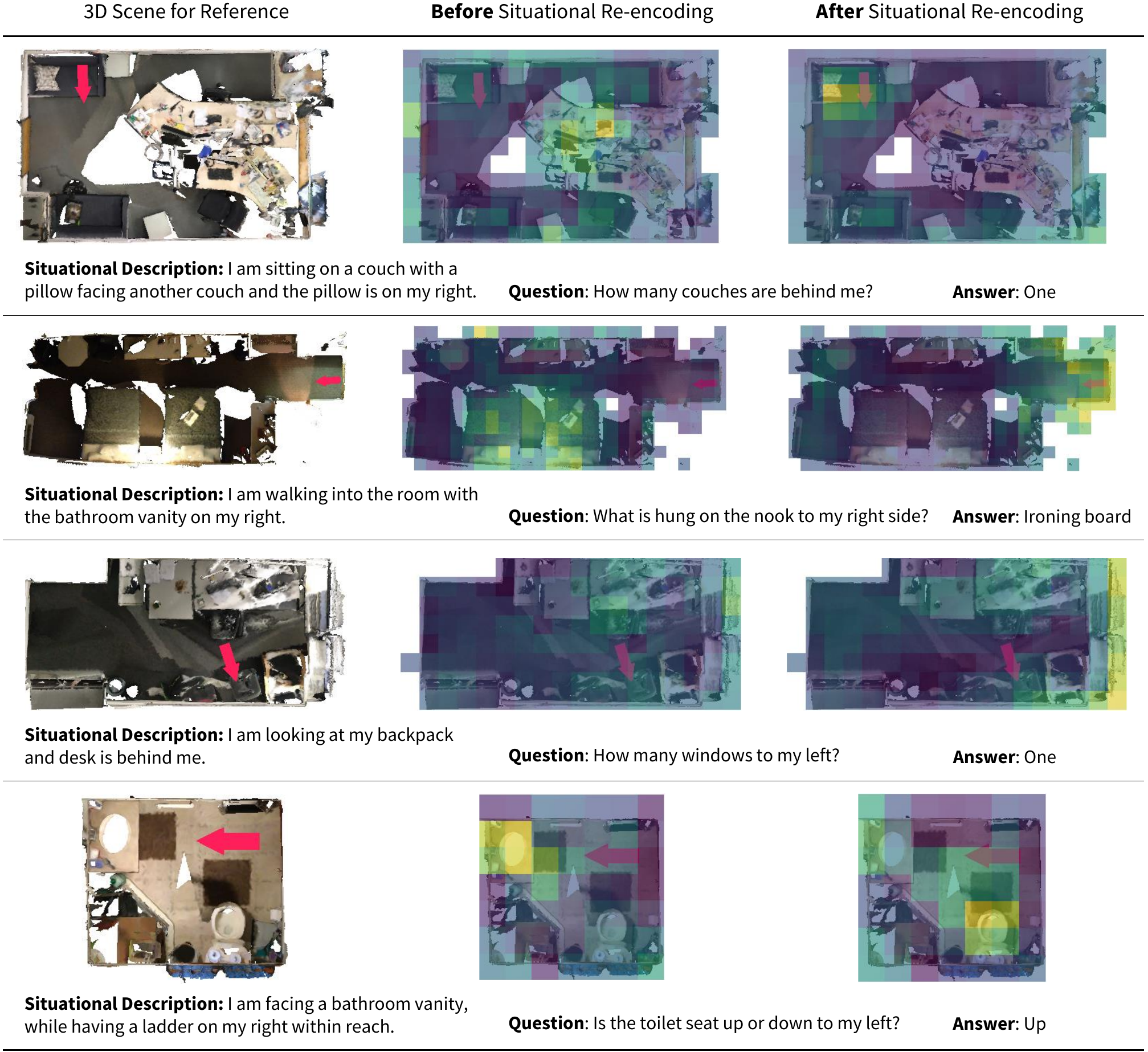}
    \vspace{-2mm}
    \caption{\textbf{3D vision token activation before and after situational re-encoding}. We can notice that higher weights are assigned to question and situation-related tokens after our proposed situational re-encoding mechanism.}
    % \vspace{-3mm}
    \label{fig:activation}
\end{figure*}

In~\Cref{fig:activation}, we provide an insightful visualization of the activation changes in 3D visual tokens $z^{\mathrm{3D}}$, before and after undergoing our situation-guided visual re-encoding process. This visualization employs the \textit{viridis} colormap, where a brighter token representation indicates a higher activation value. The effectiveness of situational guidance in amplifying the relevance of crucial tokens is evident from this depiction.

\begin{figure}[!t]
    \centering
    \vspace{-1mm}
    \includegraphics[trim=128 662 176 570, clip=True, width=\linewidth]{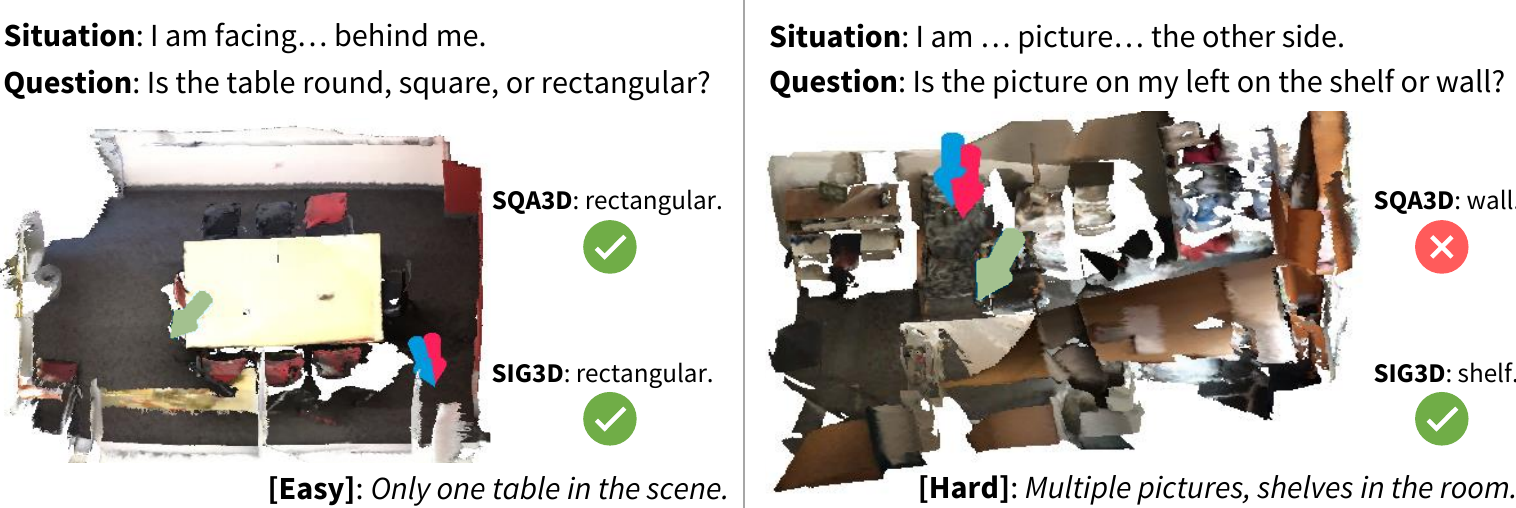}
    \vspace{-5mm}
    \caption{Visualization of cases where situation prediction benefits the most. Arrow colors: \textcolor{red}{\footnotesize GT}, \textcolor{blue}{\footnotesize SIG3D}, and \textcolor{Green}{\footnotesize SQA3D} predictions.}
    \vspace{-3mm}
    \label{fig:rebuttal-case-study}
\end{figure}

For example, the visualization in the second row reveals a notable shift in focus. Initially, the tokens predominantly concentrate on the bed area. However, after re-encoding, there is a discernible shift in attention towards areas closely aligned with the situational vector and those directly related to the query. Similarly, in the third row, the situational re-encoding process results in the window region ``on the left'' receiving increased emphasis. In the fourth row, the attention initially focuses on the vanity region. Then it shifts to the toilet on the left of the agent, as suggested by the situational vector and the question prompt. This experiment provides a clear demonstration of how our method, using enhanced situational awareness, contributes to improved performance in downstream reasoning tasks in an explainable manner. The ability of our model to dynamically adjust focus in response to situational cues is a key factor in its enhanced reasoning capabilities.

\begin{figure*}[t]
    \centering
    \includegraphics[trim=0 200 0 40, clip=True, width=\linewidth]{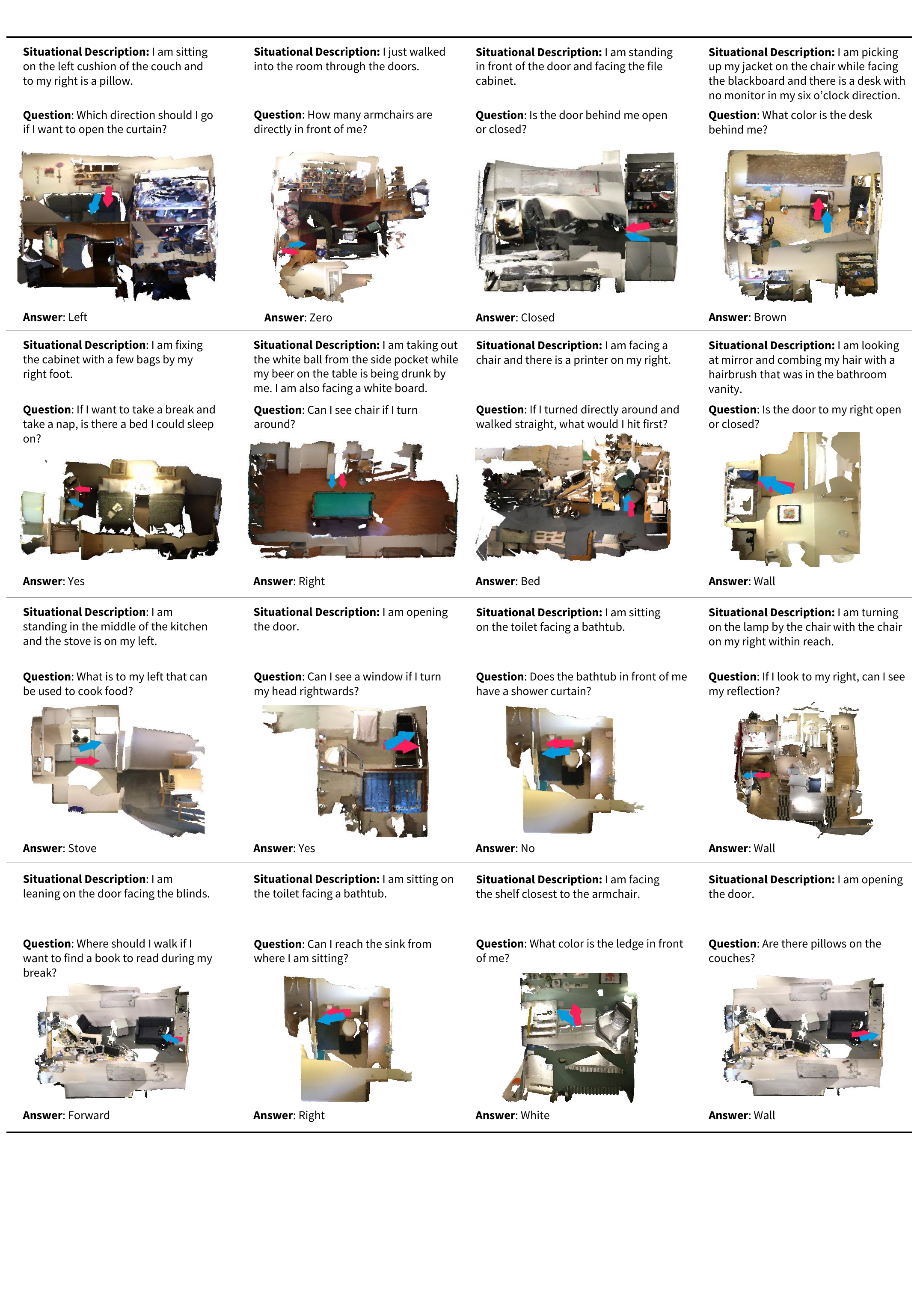}
    \vspace{-6mm}
    \caption{We demonstrate more successful examples of our method.}
    % \vspace{-3mm}
    \label{fig:qualitative_good_01}
\end{figure*}

\begin{figure*}[t]
    \centering
    \includegraphics[trim=0 200 0 60, clip=True, width=\linewidth]{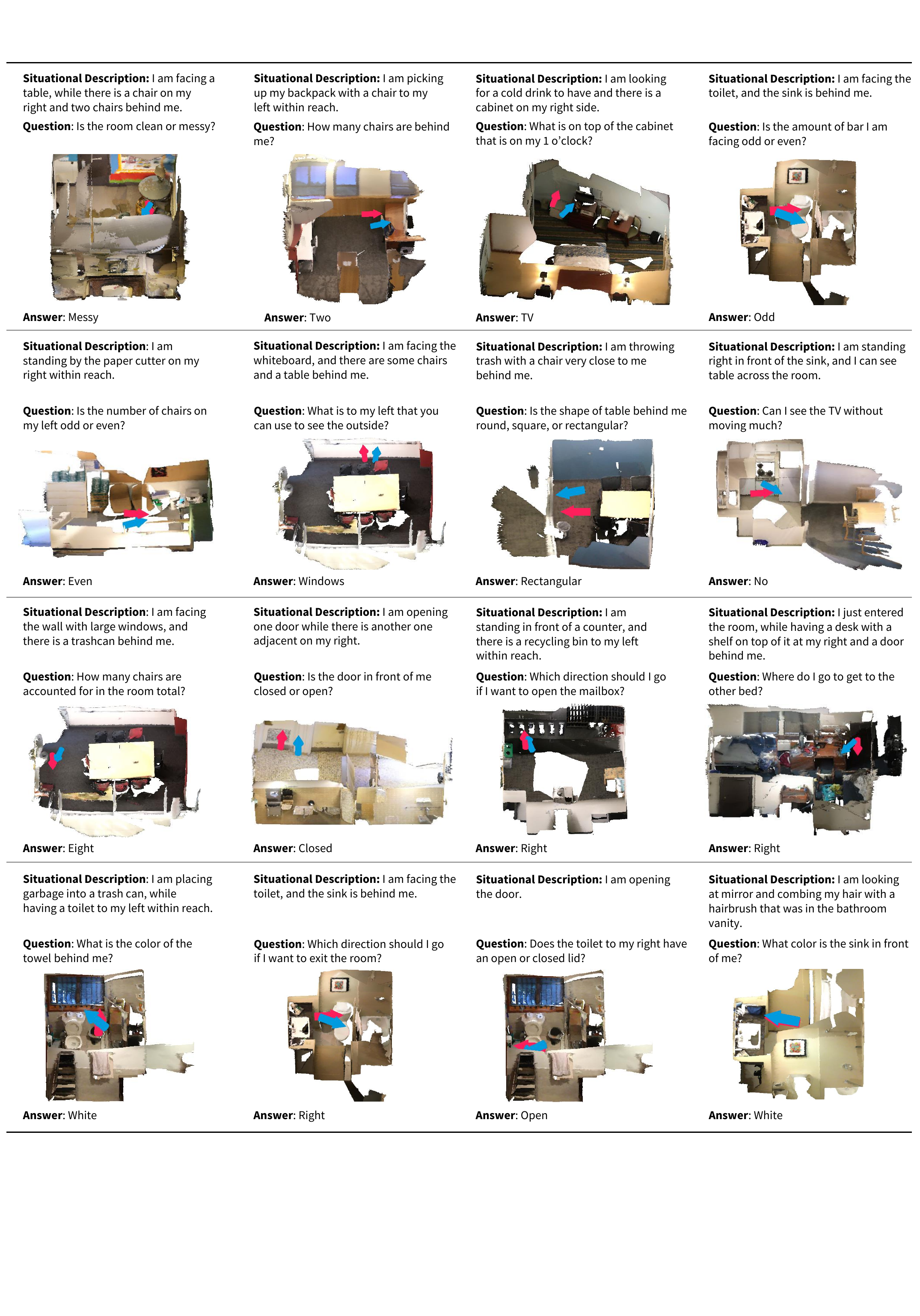}
    \vspace{-6mm}
    \caption{We demonstrate more successful examples of our method. }
    \vspace{3mm}
    \label{fig:qualitative_good_02}
\end{figure*}

\section{More Qualitative Results}\label{sec:supp_more_visualization}

We show more qualitative results of our model in~\Cref{fig:qualitative_good_01,fig:qualitative_good_02}. Visualization encompasses a diverse array of tasks, including queries about object orientation, characteristics of specific objects, the count of objects within a scene, and yes/no questions based on commonsense reasoning. A key observation from these results is that, in numerous instances, absolute precision in situation estimation is not a prerequisite for our model to accurately deduce the answers to the posed questions. This finding highlights the model's robustness and its capacity to effectively handle a variety of query types, even with less optimal situational awareness.

\section{Performance on Hard Cases}\label{sec:supp_more_challenging}

An example of our case study of easy-hard samples is shown in~\Cref{fig:rebuttal-case-study}. We find that simple examples in the dataset allow existing models to guess the correct answer without any 3D situational understanding. However, our method effectively improves the hard examples with complicated and entangled questions and situations.

\begin{figure*}[t]
    \centering
    \includegraphics[trim=125 35 135 25, clip=True, width=\linewidth]{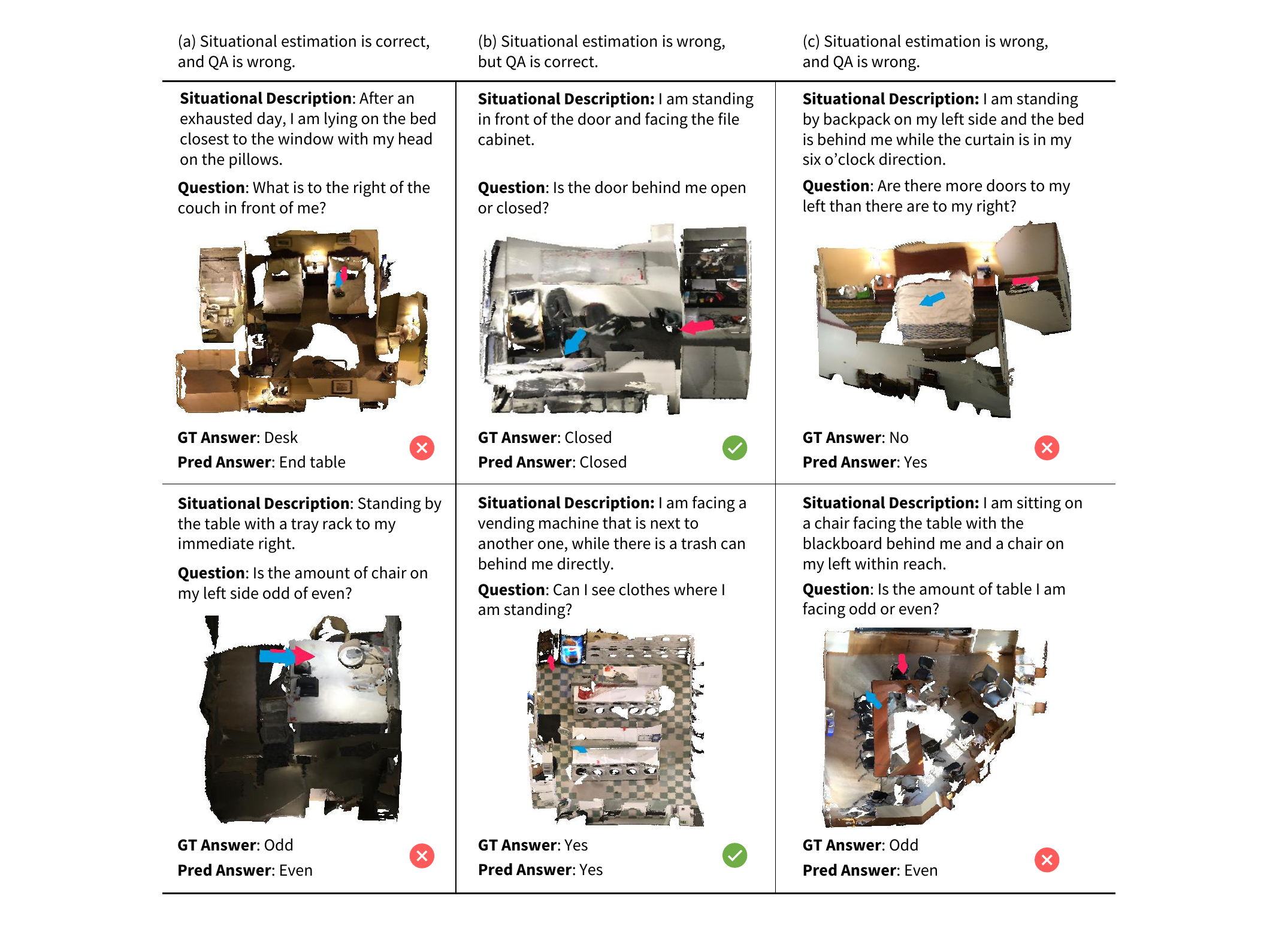}
    \vspace{-6mm}
    \caption{We demonstrate three different categories of failure cases.}
    % \vspace{-3mm}
    \label{fig:qualitative_failure}
\end{figure*}

\section{Analysis of Failure Cases} \label{sec:failure_cases}

We perform a failure case analysis on our model in~\Cref{fig:qualitative_failure}. We categorize and visualize three types of failure cases.

\vspace{1.5mm}
\mypar{Accurate Situation Estimation, Incorrect Question Answering.} This scenario demonstrates that accurate situational understanding does not necessarily guarantee correct responses to queries. A significant proportion of failures within this category can be attributed to complex question prompts that demand multi-stage reasoning or the integration of commonsense knowledge. For instance, the initial example necessitates the model's comprehension of the spatial relationship between the viewer's perspective and the couch, followed by an additional reasoning phase focused on the couch to accurately respond to the query. The subsequent example demands an understanding of the concepts of ``odd'' and ``even,'' and their application to the count of objects in a 3D environment.

\vspace{1.5mm}
\mypar{Inaccurate Situation Estimation, Correct Question Answering.} This category reveals that errors in situation estimation are more likely when the scene description involves minor or less common objects. Furthermore, it is observed that the model might incidentally arrive at the correct answer without fully grasping the complex situational and multi-modal context, particularly in cases where the question involves choosing between two or among multiple given options. Therefore, a blend of qualitative and quantitative assessments is crucial for a comprehensive evaluation of the model's performance.

\vspace{1.5mm}
\mypar{Both Situation Estimation and Question Answering are Incorrect.}This group contains the most challenging examples from the dataset, typically encompassing multiple complexities identified in the preceding categories. These cases present a compounded difficulty level, highlighting the model's limitations in scenarios that require an intricate understanding of both situational context and question interpretation.

\section{Limitations and Future Work} \label{sec:limitation_future_work}

\mypar{Selection of 3D Scenes.} The SQA3D~\cite{ma2022sqa3d} and ScanQA~\cite{azuma2022scanqa} datasets, both derived from the ScanNet~\cite{dai2017scannet} dataset, exclusively feature indoor household environments. These static scenes limit the model's applicability to dynamic tasks like manipulation and exploration. Consequently, our current model is tailored to static household settings. This scalability problem is a long-standing challenge for all existing 3D VL reasoning work~\cite{ma2022sqa3d,azuma2022scanqa,hong20233d,zhu20233d}. We believe that with a more scalable visual representation (\eg, scene graphs, sparse learnable embeddings), we can extend our model to support larger 3D environments in the future work.

\vspace{1.5mm}
\mypar{More Comprehensive Visual Encoding.} In our approach, the utilization of a voxel-based, open-vocabulary 3D encoder achieves much better overall performance. Nevertheless, for specific queries involving counting or referencing, a detection-based encoder may yield a more advantageous visual token set, owing to its capacity to provide instance-level information pertinent to the questions. This indicates the potential benefits of a multifaceted visual tokenization system that amalgamates the strengths of various encoder types.

\end{document}